\documentclass{article}
\usepackage{amsfonts}
\usepackage{amsmath}
\usepackage{graphicx}
\usepackage{graphics}

\newcommand{\R}{\mathbb{R}}

\newcommand{\argmax}{\arg\!\max}

\begin{document}
\title{Massive Data Clustering in Moderate Dimensions from the Dual Spaces of 
Observation and Attribute Data Clouds}

\author{Fionn Murtagh \\
Email: fmurtagh@acm.org}

\maketitle

\begin{abstract}
Cluster analysis of very high dimensional data can benefit from the properties 
of such high dimensionality.  Informally expressed, in this work, our focus is 
on the analogous situation when the dimensionality is moderate to small, relative
to a massively sized set of observations.  Mathematically expressed, these
are the dual spaces of observations and attributes.  The point cloud of 
observations is in attribute space, and the point cloud of attributes is in 
observation space.  In this paper, we begin by summarizing various perspectives
related to methodologies that are used in multivariate analytics.  We draw on these
to establish an efficient clustering processing pipeline, both partitioning and 
hierarchical clustering.  
 \end{abstract}

\section{Introduction}

From the next section, up to section \ref{sect15}, we summarize methodological 
perspectives.  Alternatively expressed, we briefly describe parts of analytics
processing paths, or alternative parts of such analytics processing pipelines.  

\begin{itemize}

\item
In section \ref{sect1} there is a short introduction to the notation used. 

\item 
Section \ref{lowdimranproj} covers the traditional approach of facilitating and 
expediting analysis through the forming of a reduced dimensionality accurate 
representative, and best fit, of the given data's dimensionality. 
Random projections can approximate well this processing.  

\item 
In section \ref{sect3}, there is (i) data piling that occurs in very high 
dimensions, (ii) random projections, but here shown to be very related and 
therefore this leads to interest in their aggregate or consensus.  

\item Since convergence or a sufficiency (e.g.\ of the aggregate of 
random projections) may be relevant, section \ref{sect4} just notes how 
iterative convergence can determine the principal eigenvector and eigenvalue, 
and hence the cloud's principal axis. 

\item The centring of the point clouds is one perspective on a trivial first 
eigenvector.  Another perspective can be how Correspondence Analysis provides 
a mapping of the data relating to all that differentiates the data from a null
$\chi^2$ independence statistic.  This is summarized in section \ref{CAeigen}.

\item Section \ref{sect6} describes how a hierarchic clustering, or tree 
structuring of interrelationships in the data, or a point set in an ultrametric, 
or tree, topology, can be easily mapped onto a one-dimensional alignment or ranking
or curve.  

\item While fidelity of representation is one issue with low dimensional 
mapping, in section \ref{sect7}, more at issue is the one-dimensional 
alignment or seriation that could be usable for a hierarchical clustering, or, 
otherwise expressed, an ultrametric or tree topology embedding.

\item The properties of the Baire metric, and simultaneously, ultrametric, 
are summarized in section \ref{baire}.

\item Section \ref{sect9} notes how, here, partitions are assembled into a
hierarchy, whereas traditionally, partitions are determined from a hierarchy.

\item Further refining  the hierarchical clustering that is associated with 
endowing the data with the Baire metric is considered in section \ref{sect10}.
Benefits can be: storage; and interpretability (of the clustering analytics 
being carried out).  

\item Many aspects before section \ref{sect11} relate to very high dimensional
spaces.  Given the dual space relationship, in this section interest is shifted
to massive sized data sets. 

\item One way to exploit the dual space relationships is to infer in the 
high dimensional cloud from processing carried out in the lower dimensional 
cloud. Section \ref{sect12}.

\item In section \ref{sect13}, at issue is use of data piling in one of the 
clouds, but not the other. 

\item How this all relates to seriation, the unidimensional mapping that
might serve for reading off from it, a hierarchical clustering, i.e.\ 
an ultrametric or tree topology embedding, is summarized in section \ref{sect14}.

\item Section \ref{sect15} discusses the mapping of seriation into hierarchical
clustering.   Section \ref{sect16} starts to deal with practical implementation.
Section \ref{sect17} pursues a case study in detail. 
\end{itemize}

\section{Introduction: Dual Spaces of Clouds of Points, Dimensionalities}
\label{sect1}

Clouds of points are at issue here, where the observation points cloud is 
crossed by the attribute points clouds.  Let $I$ be the index set of the 
observation cloud, $ {\cal N}(I)$, let $J$ be the index set of the attribute
cloud, ${\cal N}(J)$, and we have the observation set $x_I \subset \R^{|J|}$ and
the attribute set $x_J \subset \R^{|I|}$.  So the observation vector, $x_i$ 
is a $|J|$-valued vector.  Also the attribute vector, $x_j$ is a $|I|$-valued
vector. 

Key processing elements in the description to follow are the 
normalizing or scaling carried out, on our data, and the mappings or projections, 
that are carried out.  Both following on from the processing stages, and also 
arising from the initially obtained or sourced data, this both implies, and 
is directly related to the data distributional properties.   A further 
important distinctive property, in the description to follow, is the case when 
cardinality is very large.  This is, to begin with in the following sections,
when the ambient dimensionality of our observations is of very large 
dimensionality, i.e.\ $|J|$ is very large, and $|J| \ll |I|$.  Then we consider 
when the observation set is very large, i.e.\ $|I|$ is huge, and $|I| \ll |J|$. 

Previous applications include chemical data (high dimensional) \cite{chemical};
astronomy data (univariate, the aim being linear computational time hierarchical
clustering), \cite{conmur}; document similarity data \cite{slds}; protein 
clustering in computational biology, and enterprise information retrieval, 
 \cite{contrerasphd}.

\section{Low Dimensional Mapping from High Dimensions through Random Projection}
\label{lowdimranproj}

First, we note this aim of accurately mapping high dimensional data into a low
dimensional space.  By ``accurate'' is means preserving inter-point distances. 

Random mapping of a high dimensional cloud of points into a lower dimensional 
subspace, with very little distortion of (dis)similarity, is described in \cite{kaski}.
This aim is to consider points $g, h \in \R^\nu$, i.e.\ $\nu$-dimensional, that will be 
mapped, respectively, onto $x, y \in \R^m$, by means of a random linear transformation.
Each mapped point in $\R^m$ is transformed to be of unit norm. Sought is to have 
$ m \ll \nu$.  Kaski \cite{kaski} cites Hecht-Nielsen as follows: the number of 
{\em almost orthogonal directions} in a coordinate system, that is determined at 
random in a high dimensional space, is very much greater than the number of orthogonal 
directions. What is examined is the effect of the linear transformation, from the original
space $\R^\nu$ to the new, randomly transformed space, $\R^m$.  Distances in the 
original space, $\R^\nu$ are close to being fully preserved in the transformed 
space, $\R^m$.  A theoretical statement of this is the Johnson-Lindenstrauss Lemma 
\cite{Johnson84}.

That linear transformation, $\R^\nu \longrightarrow \R^m$ uses vectors of zero mean, 
unit standard deviation Gaussian.  Such a transformation, in \cite{Li06}, is termed 
the case of conventional random projections.  

\section{Highly Correlated Random Projections of High Dimensional Point Clouds}
\label{sect3}

Data concentration, also termed data piling, is a known characteristic of very high
dimensional spaces.  By the central limit theorem, and by the concentration (data 
piling) effect of high dimensions \cite{hall,terada}, we have as dimension 
$\rightarrow \infty$: pairwise distances become equidistant; orientation tends to be 
uniformly distributed.  
Random projections are here carried out using uniformly, $[0,1]$, distributed random
axis coordinates.  The projections are rescaled to the (closed/open) unit interval, 
$ \geq 0, < 1$.   We find high correlation between any pair of our random projections.
See \cite{slds}.  Unlike the conventional random mapping into a low dimensional
space, as described in section \ref{lowdimranproj}, here we are using a uniform
distributed mapping rather than a Gaussian distributed mapping.  Also different from 
the conventional random mapping, projections are rescaled to the 0,1 interval, 
compared to normalizing to unit norm.  

In the conventional random mapping, the aim is approximating distance preservation
(termed the Parseval relation) in the low dimensional mapping.  Rather than this, 
with our highly correlated random mappings, we seek instead to exploit and draw benefit
from this mapping of clouds that are subject to data piling.  

In order to do this, we form the consensus random projection.  This is constituted
by the mean of each point's random projected values.  

\section{Iterative Determining of the Principal Eigenvector}
\label{sect4}

In \cite{ecda2015}, there is consideration of the iterative determination of 
the principal eigenvector.  Such an algorithm is easily and straightforwardly 
implemented.  (However it is computationally less interesting and relevant, 
compared to eigen-reduction through diagonalization.)   Also considered in 
\cite{ecda2015} is power iteration clustering, basing the clustering on the 
principal axis projections.  

There is discussion in \cite{jpb1982} of such iterative eigenreduction, with an 
unbounded observations set, hence indicated as an infinite number of observations 
or input data table rows.  

\section{Correspondence Analysis Decomposition: Trivial Eigenvectors}
\label{CAeigen}

In the eigen-decomposition, we have the following expressing, for $N = |I|$, 
and for factors, $F, G$.  

\begin{equation}
f_{ij} = f_i f_j \left( 1 + \sum_{\alpha = 1, \dots N} \lambda_{\alpha}^{-\frac{1}{2}}
F_{\alpha(i)} G_{\alpha(j)} \right) 
\label{eqn1}
\end{equation}

Here the first term on the right gives rise in practice to trivial eigenvalues, 
equal to 1, and to associated trivial eigenvectors.  If we have, through 
data piling, a very large number of factors, in the summation, then it comes 
about that there is an approximation of the data, $f_{ij}$ by $f_i f_j$.  
This, therefore, is the product of the marginal distributions.  This, in turn, 
is given by a constant times the row and the column sums.

It may be just noted, that this amounts to a 0-valued $\chi^2$ statistic, that
fully supports expected values, defined from the row and column margins, 
specifying the observed data.  With data piling, or data compactification, we
can, and will, arrive at such an outcome. 

\section{Hierarchical Clustering and Seriation}
\label{sect6}

A dendrogram's terminal nodes are a permutation of the object set that is 
clustered.  Let us denote that permutation of the object (or observation) set, 
$I$, as $\pi(I)$.   While of course the pairwise distances defined on $I$ 
determines the hierarchical clustering, it is also very clear that that many 
variants on these pairwise distances would also determine an identical 
hierarchical clustering.  Thus, one important characterizing aspect of a 
hierarchical clustering is $\pi(I)$, and its inter-point adjacent distances 
that lead to the hierarchical clustering.  In \cite{critchley}, there is the 
foundation for hierarchical clustering requiring a seriation of what is to be
clustered.  

It is this seriation therefore, that we use, based on the consensus, that is, 
the highly correlated, set of random axis projections resulting from data 
piling, or compactification.  

\section{Two Distinct Objectives of Low Dimensionality Embedding}
\label{sect7}

Compared to section \ref{lowdimranproj} where the aim was the approximating of
pairwise distance invariance in a low dimensional space embedding, here we bypass
this objective in our analytics.  From one perspective, such distance invariance
is of benefit if we wish to carry out best match or nearest neighbour processing, 
with the approximation of every pairwise distance in the original space, and
in the embedded space.  

Arising out of how (see \cite{critchley}) what is hierarchically clustered 
can be perfectly 
scaled in one dimension, we want to use this knowledge to proceed directly to our
clusters.  These clusters, being hierarchically structured, are determined on 
most, if not all, resolution scales.  That is due to the taking full account of 
cluster embedding, or cluster inclusion, properties.   

Here, therefore, our overriding objective is not so much fidelity in
a new data space relative to our original data.  Rather, we are concerned with 
both effective and efficient data scaling.  

One form of realising such an objective is as follows.  Our given data clouds are 
endowed with an appropriate metric, such as for positive counts or other real 
measurements, the $\chi^2$ metric. That is, the data clouds are
endowed with the Euclidean metric.  Distance invariance will hold between (i) 
$\chi^2$ distance in input space, and (ii) Euclidean distance in the output, factor 
space.  That output space, by design, is also referred to as comprising: 
factors, principal components, principal axes, eigenvectors.

\section{Baire or Longest Common Prefix Distance}
\label{baire}

Consider the consensus, viz.\ aggregated set of projections.  Given two values, 
$x_{ik}, x_{i' k}$ for digits $k = 1, 2, ... , K$, the Baire distance of base $B$
is defined as $B^s$ where $s = \argmax_k x_{ik} = x_{i' k}$.  

In view of potential 
use of different number systems, $B$ should be the number base that is in use, 
e.g.\ $B = 10$ for decimal and real numbers. 

See \cite{conmur} for the effective distance such that cluster members can be 
directly read off the data, once we specify each cluster, as a bin, specified 
by the set of shared, common prefix values.  

The Baire, or longest common prefix distance, is both a metric and an ultrametric. 

A property of the use of this metric, and ultrametric, is that we can specify any 
cluster, using the common prefix.  Then we can read off the cluster members from 
our data.  That is, a linear scan of the data is carried out to determine all 
members of a given cluster, specified by what is to be the common prefix.  Thus, 
by endowing one's data with this metric and ultrametric, there is direct reading of
the cluster members.  

\section{Practical Strategies for Hierarchical Clustering}
\label{sect9}

Traditionally, one use of agglomerative hierarchical clustering has been to determine, 
for $n$ observations, the extraction of one of the $n-1$ partitions that are defined by 
the hierarchy.  Motivation for that is lack of prior knowledge of number of clusters
in a targeted partition.    

Given the direct reading of clusters that are hierarchically structured, for very 
large datasets, it is beneficial to also extract one or more partitions.  These 
partitions are on a range of resolution scales.  

Taking our hierarchy as a regular $B$-way tree, for example with $B = 10$, then the
first level of the hierarchy has a partition with $B$ clusters, the next level has 
a partition with $B^2$ clusters, and so on.  In practice, it is a data-dependent 
issue as to whether any of these clusters are empty.  

\section{Sparse Encoding of the Hierarchical Clustering}
\label{sect10}

The regular $B$-way tree may be refined, using linear computational time processing, 
in order to regular $B-1$-way tree, $B-2$-way tree, and continuing.  The aim, in 
this stepwise refinement, to exploit sparsity in the data.  

It is interesting to find empirically that p-adic encoding, p prime, may best represent, 
i.e.\ approximate, the given hierarchical tree, \cite{sparsepadic}.

\section{Optimal Dimensionality Reduction of the Dual Cloud Spaces}
\label{sect11}

Just to begin, consider our point cloud of observations in the attribute space, 
and, equivalently, our attribute point cloud in the observations space.  
Statistically optimal reduction of dimensionality, in Principal Component Analysis (PCA)
for example, linearly transforms the point cloud exploiting those attributes that 
best preserve the variance of the point cloud.  Such eigenreduction is of cubic 
computational complexity.  Due to the dual space relationship, the determining of
eigenvectors providing the latent, principal axes or factors, and the associated
eigenvalues providing the variance explained by those axes, is carried out in 
either of our dual spaces.  That is, carried out in either of the point clouds.

\section{Clustering of a Point Cloud Inferred from Its Dual Space}
\label{sect12}

Given the inherent relationships between the dual spaces, clustering of one cloud
can be used to infer clusters in the dual space.  Thus, we might consider clusters 
of attributes leading to clusters of their associated observations.  

This can be considered as the basis for block mode clustering, see \cite{liiv}.
For statistical inference of clusters in a dual space, see also the {\tt FactoMiner}
package in R.

\section{Exploiting Dual Space Relationships for Either $|I| \ll |J|$ or Vice 
Versa}
\label{sect13}

Previous sections have considered how a seriation can be constructed whenever 
data piling arises, through massification of data.  Informally, this might be 
expressed as follows.  Data piling is compactification, or becoming condensed.
The new origin of this mapped, or embedded, data is located at the centre of 
this piled or compactified data.  The rescaling that is used will ensure that
there is a point norm-based ordering.  

Subject to setting up the data, as described in section \ref{CAeigen}, there 
will be this ``condensation'' of the data clouds resulting in: $f_{IJ} = 
f_I f_J$.   When one of these clouds becomes very much compactified, that data piling
can be viewed as approximating a single massive point.  Geometrically this can be 
considered as becoming the product of a scalar and a vector.  This therefore points 
to the relevance of one of the marginal distributions as defining the unidimensional
mapping that is sought, to be the seriation to be used for the clustering.  

\section{Summarizing These Approaches to Dual Cloud Embedding, to Derive a 
Seriation of the Observations} 
\label{sect14}

Previous sections can be empirically verified.  Key elements include: relative 
very large dimensionality; the normalization and rescaling that are used on the 
data prior to mapping to a Euclidean metric endowed space; normalization or 
data recoding that is used in the mapping.

One further aspect of importance is are distributional characteristics of the 
data, at all stages of the processing.  This has a practical aspect also, in
that applicability to massive data volumes gives rise to the need to 
appropriately encode one's data.  Part and parcel of data encoding is the resolution 
scale of the data.  Our use of hierarchical clustering is strongly motivated by this 
requirement for practical adaptability in applications. 

It may be therefore necessary in practical applications to check on, and to monitor, 
data distributional properties, in order to benefit if distributional configurations
are particularly simple, or if distributional configurations have problematic 
features. 

\section{From Seriation to Hierarchical Clustering}
\label{sect15}

Hierarchical clustering through direct reading, for a given resolution level or 
for a given partition, can be viewed as quantization of the distribution of 
mapped observations onto the seriation structure.  Traditional clustering, 
model-based or k-means, involves optimal quantization with non-fixed
thresholds.  Then in non-uniform quantization, with each label
we associate a codebook entry, or associated cluster mean.  See \cite{quantization}
for extensive discussion.  

The regular hierarchical tree that results from the Baire metric and ultrametric
provides quantization that has non-fixed thresholds relative to the clusters 
that are formed.  

The Baire hierarachy or tree that results is an effective and efficient clustering
method; it is adaptive; it is highly adaptable and adaptive for massive data sets;
unlike traditional hierarchical clustering which may have partitions derived from it,
in effect the Baire hierarchy or tree is constructed from the succession, in a top-down
manner, of partitions.

As noted following section \ref{lowdimranproj}, precision relative to 
measured interpoint distances is not our objective.  In a very massive data context, 
such precision is not a primary interest.  Instead, let the following informal 
statement be considered.  In astronomy, that deals with observational data from the 
cosmos, it is established and usual practice to see objects and clusters of 
objects that are deemed to be candidates, that is, candidates for selective, 
detailed or close-up further analytics.  

\section*{Further Examples of Clustering through Quantization.}

We conclude with a further commentary on quantization as an approach to 
clustering.  In \cite{ecda2015}, there is the following.  

Using our approach on the Fisher iris data, \cite{fisher}, 150 flowers crossed by petal
and sepal width and breadth, provides the following outcome.  We determine row sums, of
the initial $150 \times 4$ data matrix, and the mean random projection of projections
on 100
uniformly generated axes.  From our previous results, we know that these are very highly
correlated.  We construct hierarchical clusterings on (i) the original $150 \times 4$
data matrix, (ii) the mean random projection, and (iii) the row sums.  The cophenetic
correlation
coefficient is determined.  (This uses ultrametric distances derived from the hierarchical
tree, or dendrogram.)   We find the cophenetic correlation of the hierarchies constructed
on the row sums, and on the mean random projection, to be equal to 1 (as anticipated).
Then between the hierarchy constructed on the $150 \times 4$ data matrix, and the
mean random projection, the cophenetic correlation coefficient is 0.8798.  For the given
data and the row sums, it is 0.9885.  The hierarchical clustering used was the average
method; and other methods, including single link, provided very similar results.  The
distance used, as input to the hierarchical agglomerative clustering, was the square root
of the squared Euclidean distance. Other alternatives were looked at, from the point of
view of the distance used, and from the point of view of the agglomerative hierarchical
clustering criterion.

We also looked at uniformly distributed, on [0,1], data of dimensions $2500 \times 12$.
The correlation between row sums and mean of 100 random projections was 0.99.  However,
for the correlation between the hierarchical clustering on the original data, and the
mean random projection, this correlation
was 0.58.  The correlation with the row sums was 0.578.  The
performance on this randomly generated data is seen to be not as good as that on the
real valued, Fisher data.  For data which is not strongly clustered, quantization
is relevant.  In the k-means clustering (partitioning) context, see e.g.\
\cite{lloyd}.  Descriptively expressed, in quantization, in addition to cluster
compactness, approximating identical cluster sizes is an objective.

\section{Practical Case Studies of Mapping Seriation to Hierarchy}
\label{sect16}

In \cite{endm}, it is shown on Fisher's iris data how hierarchical clustering 
using a seriation that is the aggregate or consensus random projection, and the 
row mass distribution, is well and truly associated (high cophenetic correlation,
that is, correlation of the ultrametric distances) with an agglomerative 
hierarchical clustering method. 

\section{Case Study, Towards Both Partition-Based Clustering 
and Hierarchical Clustering with Linear Computational Complexity}
\label{sect17}

We draw on the wide-ranging vantage points offered by the methodologies for 
carrying out clustering, and related analytics processing.  The latter may 
possibly include: dimensionality reduction, orthonormal factor or principal 
axis space mapping, feature or attribute selection, and so on.  Effectively we
are drawing on what has been overviewed in previous parts of this paper.  

\subsection{Step 1: Determining the Seriation}

Gene expression data are used, where rows contain genes and columns contain
samples, example in Table \ref{tab1}. 

\begin{table}
\begin{tabular}{lrrrrrr}\hline
gene-id  & GSM177577 & GSM177578 & GSM177579 & GSM177580 & GSM177581 & GSM177582 \\ \hline
U48705.01    &  4091.2  & 3683.3 & 3117.9 & 3775.1 & 3510.3 & 3461.4 \\
M87338.02    &  762.7  & 666.9  & 581.2 &  623.5 &  822.9  & 735.9 \\
X51757.03    &  113.3  & 112.8  & 90.5  &  128   &  120.7  & 93.1 \\ \hline
\end{tabular}
\caption{A sample of the data. Genes crossed by samples.}
\label{tab1}
\end{table}

The genes dataset used is a table of dimensions $61359 \times 16$.
The data, $x_{ij}$ values for $i \in I, |I| = 61359, j \in J, |J| = 16$, 
are extremely exponentially distributed in their values; the row sums, $x_I$ such 
that $x_i = \sum_{j \in J} x_{ij}$, are also extremely exponentially distributed;
and the column sums, $x_J$ such that $x_J = \sum_{i \in I} x_{ij}$, is Gaussian.
For the latter, the Shapiro-Wilk normality test gives a p-value of 0.9145.  
Just to illustrate values of $x_{IJ} = x_{ij} \forall i \in I, j \in J$ we ave
the maximum, minimum, mean and median as follows: 676496, 0.199789, 1130.097, 130.5835.

Next checking the mean of random projections, with uniformly distributed axis 
coordinate values, we find very high correlation with the row masses ($f_I = 
x_I / \sum_{i, j} x_{ij}$), equivalent for correlation to the row sums, $x_I$.
Note the ordinate scale, that starts, with just one random projection, at a 
correlation of almost 0.999.  

\begin{figure}
\centering
\includegraphics[width=8cm]{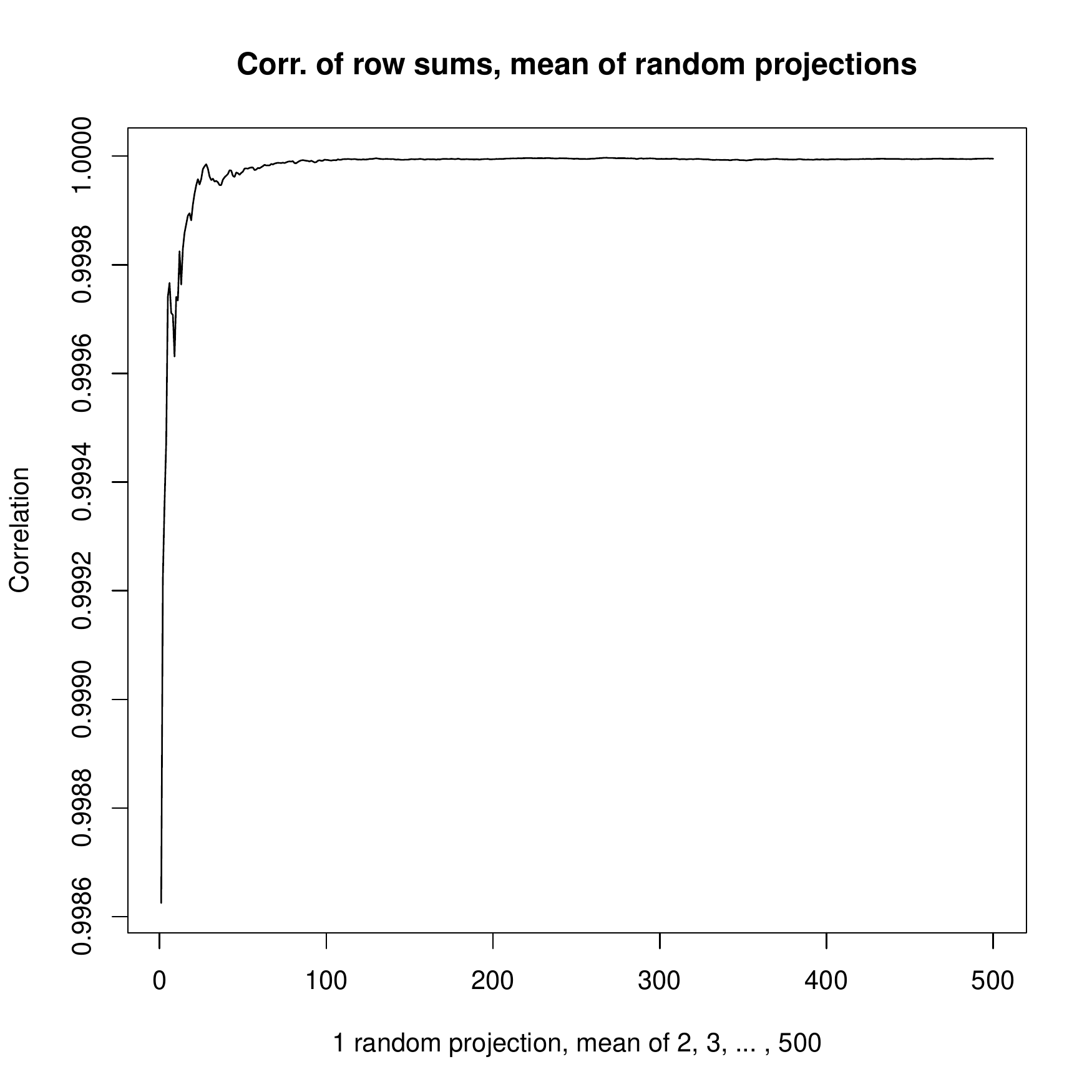}
\caption{Correlation between the row sums, and the mean random projection.  The 
random projections are the 1st, then the mean of the 1st and 2nd, the mean of the
1st, 2nd, and 3rd; and so on, up to the mean of the 1st, 2nd, 3rd, 4th, ..., 500th
random projections.}
\label{fig1}
\end{figure}

This first finding indicates how very consistent the scaling is, when taking 
the seriation of genes, that is provided by any of the following: (i) distributional
masses of the genes (notationally here, $f_I$). 

\subsection{Step 2: Transforming or Re-Encoding the Seriation}

Arising out of the previous subsection, exemplifying the simplification property of 
cloud one of the clouds ${\cal N}(I)$, ${\cal N}(J)$ stated in \ref{CAeigen}, 
we proceed to examining $f_I = \{ f_i | i \in I \} $ where $f_i = \sum_{j \in J} x_{ij} 
/ \sum_{i \in I, j \in J} x_{ij}$. 

The seriation is to be directly used for deriving a partition of clusters or a hierarchy
of clusters.  This is to be carried out with computational time that is 
linear or $O(n)$ for $n$ observations, using earlier notation, $n = |I|$.

The distribution of the seriation values is found to be highly exponentially decreasing.  
A direct, density-based, determining of clusters could lead to just one single 
cluster with all observations as members of it, but this would be futile in 
practice.  Hence we would like to re-encode our seriation values to carry out 
quantization-based determining of clusters.  Cf.\ section \ref{sect15}.  We could 
sort our seriation values and then read the values, for a top-level partition with 
10 clusters, as a series of 10\%, 20\%, etc.\ quantiles.  However, sorting the 
$n$-length set of seriation values requires $O(n \log n)$ computational complexity. 
We see rather to bound our processing to be $O(n)$.  

Due to exponentially distributed values, we first take the log of these values.
Since this is Gaussian distributed, we standardize it to zero mean and unit 
standard deviation.  

Next we uniformize the standard Gaussian distributed values, i.e.\ we convert to 
a uniform distribution, using the complementary error function.  
Complementary error function is one minus the error function; the error function is
the probability of a value being within the range $0, x/\sigma \sqrt{2}$  The error
function is twice the integral of a normalized Gaussian function in this interval. 
Cf.\ \cite{culham}.  

\begin{figure}
\centering
\includegraphics[width=8cm]{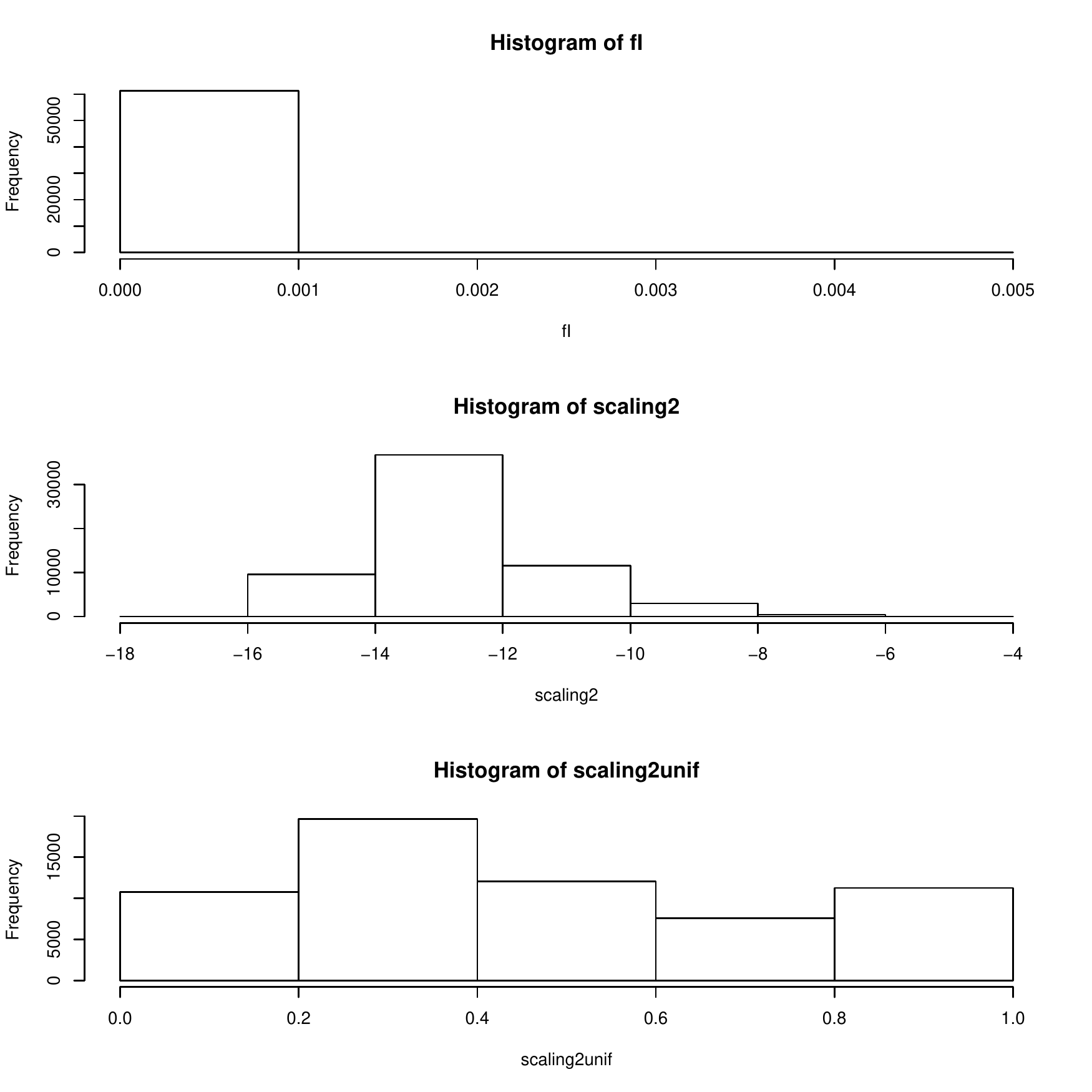}
\caption{Histograms of marginal distribution; of standard normal distributed 
re-encoding; and of uniformized re-encoding.}
\label{fig2}
\end{figure}

The effectiveness of these re-encodings is displayed in the histograms in 
Figure \ref{fig2}.  Our aim is to have a reasonable strategy that is computationally
of linear complexity, for taking the highly exponentially distributed values into 
a near uniformly distributed set of values.  By a reasonable strategy, we mean 
an approach which is generalizable, which is assessed at least visually using 
histograms, and also which can be further considered or indeed, subject to 
alternative generalizable and computationally scalable approaches.  

\subsection{Step 3: Reading off Partitions at Successive Hierarchical Levels}

We directly read off the cluster members by taking the first, viz.\ top, 
partition as consisting of 10 clusters.  Our uniformized data values are in 
the closed/open interval, $[0,1)$.  We find the following cluster cardinalities,
respectively for the clusters that are labelled 0, 1, 2, 3, 4, 5, 6, 7, 8, 9: 
2345, 8445, 10347, 9316, 6948, 5122, 3999, 3588, 3926, 7323.  

Equally straightforwardly we can read off the next partition in a regular 
10-way hierarchy or tree, that define the clusters with labels 0, 1, 2, $\dots$, 
98, 99.  Following the top partition with 10 clusters, that second level 
partition with 100 clusters, then the next, third level, partition would have 
1000 clusters, and so on.  

In the Appendix, the R code used is listed.  

\subsection{Cluster Properties of Top Level Partition}

In order to validate the clustering, here the properties of the top level 
partition, with 10 clusters, are examined.  The distribution of the given 
data, as noted above, is exponentially decreasing.  We determine the mean vectors 
of the clusters in the given 16-dimensional attribute space.  Next, all pairwise
distances are obtained.  These are Euclidean distances.  For each cluster, 
the nearest neighbour cluster is determined.  Because of the distributional 
properties of our data, we find that cluster 0 has as its nearest neighbour, 
cluster 1; cluster 1 has as its nearest neighbour, cluster 2; and so on, in 
succession.  These nearest neighbour clusters, and the nearest neighbour 
distances, are listed in Table \ref{tab2}.   

To summarize how compact the clusters are, for each cluster we determine the 
variances in each dimension, aggregate (sum) these variances to have the 
overall cluster variance, and then list the overall cluster standard deviation, 
just multiplied by 3, in Table \ref{tab2}.  The $3 \sigma$, i.e.\ 3 
standard deviation, measure used is just so as to have comparability with 
what would be relevant for a Gaussian distributed cluster. 

What Table \ref{tab2} demonstrates very well, is that cluster compactness, 
measured by the $3 \sigma$ column, is far less in value than the cluster-to-cluster 
nearest neighbour distance.  Therefore this leads to the conclusion 
that the cluster properties, in this top level partition into 10 clusters, 
are excellent.  

\begin{table}
\begin{tabular}{crcr} \hline
Clusters & 3 * Std.dev & NN-cluster & NN-dist. \\ \hline
0   &  49.92530   &  1    &    94.48774  \\
1   &  85.11277   &  2    &    125.75401 \\
2   &  124.10223  &  3    &    160.40965 \\
3   &  130.90670  &  4    &    213.86643 \\
4   &  161.75631  &  5    &    296.95453 \\
5   &  205.36007  &  6    &    445.92892 \\
6   &  180.60822  &  7    &    767.96664 \\
7   &  275.07759  &  8    &    1586.48549 \\
8   &  367.02603  &  9    &    27694.34381 \\
9   &    1152.21181  &      &              \\ \hline
\end{tabular}
\caption{The top or first level partition, with 10 clusters.
Column 2 lists $3 \times$ standard deviation of the cluster.
Column 3 lists the nearest neighbour cluster of each cluster.
Finally column 4 lists that nearest neighbour distance.}
\label{tab2}
\end{table}

\section{Conclusion}

The clustering approach used here is quite simple to implement, and its 
methodological basis has been straightforwardly and briefly described in 
previous sections.  In section \ref{sect15} in particular, there is the 
brief description of the methodological foundations for this approach to 
hierarchical clustering and determination of one or more partitions.  

While it may be noted that instead of a compactness clustering criterion, 
we are using quantization as the main basis for the clustering.  
For candidate object selection, and so on, as is common in observational 
science, then this can be claimed to be quite adequate, as a methodology.  
If there are physical laws, however, at issue, then these may be used, e.g.\
through feature selection, or through statistical modelling.  For the latter,
we have noted throughout this work just what distributional properties were
found, or were verified, to hold.  

While this work has considered the data properties, it is to be considered 
therefore that this particular analytics processing pipeline, that has 
been the focus of this work, is associated with data of a given class or 
family of distributional and related properties. 

A major justification for our extensive description of methodological 
underpinnings is to allow for alternative, but related, approaches to be 
designed and implemented when there are somewhat different ultimate 
objectives.  This might include the incorporation of supervised 
learning phases in the analytics chain or pipeline.

\section*{Appendix}

In the following R code, we show dimensions and other data properties, such as
maximum and minimum values, and mean and median, in order to both indicate
data values, and to support reproducibility. 

\begin{verbatim}
> x <- read.table("dataset-1.tsv", header = TRUE, row.names=1); dim(x) # 61359 x 16
> x2 <- as.matrix(x)  # As read, x is of list data type.
> max(x2); min(x2); mean(x2); median(x2)  # 676496 0.199789 1130.097 130.5835

# Determine the marginal distribution; then re-encode from its very exponential
# distribution. 
> fI <- apply(x2, 1, sum)/sum(x2)
> min(fI); max(fI)  #  1.049568e-07 0.004014821
# Exp to normal, then normal to uniform:
> scaling2 <- log(fI)
> scaling2norm01 <- (scaling2 - mean(scaling2))/sd(scaling2)
# Next use complementary error function.  Cf. help(Normal)
> erfc <- function(x) 2 * pnorm(x * sqrt(2), lower = FALSE)
> scaling2unif <- 0.5*erfc(- scaling2norm01/sqrt(2))
> max(scaling2unif); min(scaling2unif)  #  0.9999998  0.01001248

# Top level partition with 10 clusters labelled 0, 1, ... , 9.
> scaling2class10 <- trunc(scaling2unif * 10)
# Next level partition with 100 clusters labelled 0, 1, ... , 99.
# Note that we find the cluster with label 0 to be empty.
> scaling2class100 <- trunc(scaling2unif * 100)
> scaling2class1000 <- trunc(scaling2unif * 1000) # 3rd level partition.

# Cluster cardinalities:
> table(scaling2class10)  # Top level partition, 10 clusters. 
scaling2class10
    0     1     2     3     4     5     6     7     8     9
 2345  8445 10347  9316  6948  5122  3999  3588  3926  7323
> table(scaling2class100)  # Second level partition, <= 100 clusters. 
> table(scaling2class1000)  # Third level partition, <= 1000 clusters. 
> length(table(scaling2class100))  # 99 (non-empty clusters)
> length(table(scaling2class1000))  # 989 (non-empty clusters)

\end{verbatim}

\end{document}